\documentclass[runningheads]{llncs}

\usepackage{eccv}

\usepackage{eccvabbrv}

\usepackage{amsthm}
\usepackage{amsmath}
\usepackage{graphicx}
\usepackage{booktabs}
\usepackage{times}
\usepackage{soul}
\usepackage{url}
\usepackage[utf8]{inputenc}
\usepackage{caption}
\usepackage{booktabs}

\usepackage{epsfig}
\usepackage{amssymb}
\usepackage{booktabs}
\usepackage{makecell}
\usepackage{float}
\usepackage{subcaption}
\usepackage{wrapfig}
\usepackage{multirow}

\usepackage[linesnumbered,lined,ruled]{algorithm2e}

\usepackage[pagebackref,breaklinks,colorlinks,citecolor=eccvblue]{hyperref}
\usepackage{orcidlink}

\begin{document}

\title{Personalized Federated Domain-Incremental Learning\\ based on Adaptive Knowledge Matching} 

\titlerunning{Federated Domain-Incremental Learning}

\author{Yichen Li\inst{1}\orcidlink{0009-0009-8630-2504} \and
Wenchao Xu\inst{2}\orcidlink{1111-2222-3333-4444} \and
Haozhao Wang\inst{1}\orcidlink{0000-0002-7591-5315}\thanks{Haozhao Wang, Yining Qi and Ruixuan Li are corresponding authors.}
\and \\
Yining Qi\inst{1}\orcidlink{0009-0001-3685-5989}
\and
Jingcai Guo\inst{2}\orcidlink{0000-0002-0449-4525}\and
Ruixuan Li\inst{1}\orcidlink{0000-0002-7791-5511}
}

\authorrunning{Yichen Li, et al.}

\institute{School of Computer Science and Technology, Huazhong University of Science and Technology, Wuhan, China\\
\email{\{ycli0204,hz\_wang,qiyining,rxli\}}@hust.edu.cn \and
Department of Computing, The Hong Kong Polytechnic University, Hongkong, China
\email{\{wenchao.xu,jc-jingcai.guo\}}@polyu.edu.hk}

\maketitle

\begin{abstract}
This paper focuses on Federated Domain-Incremental Learning (FDIL) where each client continues to learn incremental tasks where their domain shifts from each other.
We propose a novel adaptive knowledge matching-based personalized FDIL approach (pFedDIL) which allows each client to alternatively utilize appropriate incremental task learning strategy on the correlation with the knowledge from previous tasks. More specifically, when a new task arrives, each client first calculates its local correlations with previous tasks.
Then, the client can choose to adopt a new initial model or a previous model with similar knowledge to train the new task and simultaneously migrate knowledge from previous tasks based on these correlations.  
Furthermore, to identify the correlations between the new task and previous tasks for each client, we separately employ an auxiliary classifier to each target classification model and propose sharing partial parameters between the target classification model and the auxiliary classifier to condense model parameters. We conduct extensive experiments on several datasets of which results demonstrate that pFedDIL outperforms state-of-the-art methods by up to 14.35\% in terms of average accuracy of all tasks.   
\end{abstract}

\section{Introduction}
Federated Learning (FL) has become a basic paradigm that facilitates collaborative training of deep neural networks (DNNs) from distributed clients while preserving their data privacy\cite{mcmahan2017communication,zhang2021survey}. 
To deploy FL in real-world applications, it is practical that each client can continuously learn new incremental tasks while retaining the accuracy for the previous tasks~\cite{yoon2021federated}. 
However, history data may often be out of reach with privacy constraints like GDPR and many studies have shown that DNNs will forget the knowledge of previous tasks when learning new tasks~\cite{van2022three,tankard2016gdpr}. Such a phenomenon is known as 'Catastrophic Forgetting', which is a major challenge of incremental learning in the centralized setting. Especially, this challenge is even more formidable to solve in the FL scenario, where the limited previous data are stored in distributed clients and cannot be accessed by each other~\cite{zhang2023target,bakman2023federated}. 

Many recent studies have been proposed to solve the catastrophic forgetting challenge in federated incremental learning. For instance, \cite{yoon2021federated} aims to learn a personalized model for each client by decomposing the model parameters into shared parameters and adaptive parameters and transferring common knowledge between similar tasks across clients. FCIL \cite{dong2022federated} addresses the federated class-incremental learning and trains a global model by computing additional class-imbalance losses. Authors in \cite{ijcai2022p0303} utilize extra data at both the server and client sides to conduct knowledge distillation to alleviate forgetting. \cite{qi2023better} optimize the generative network and reconstruct previous samples for replay. Recently, Re-Fed is proposed in \cite{Li_2024_CVPR} to prevent catastrophic forgetting with synergistic replay by caching important samples which contribute to both the local and global understanding.

While these approaches have achieved great success, they are generally designed for the scenario where the new tasks are incremental classes, i.e., Federated Class-Incremental Learning (FCIL). These methods may lose their effectiveness when the new tasks have the same classes but their domain shift with different feature distributions, i.e., Federated Domain-Incremental Learning (FDIL). 
To elaborate further, the FCIL\cite{dong2022federated} method relies on changes in the new data category label in incremental tasks to alleviate forgetting of previous tasks; FOT\cite{bakman2023federated} projects different incremental tasks into different orthogonal subspaces, which means feature isolation for different tasks, but if suffers from the intersected feature space boundaries of incremental tasks. The primary challenge in implementing FCIL methods in the context of FDIL is that FCIL methods typically depend on the connection between previous and new classes, which is not applicable in FDIL scenarios where the classes remain consistent. 

In this paper, we seek to tackle the challenge of catastrophic forgetting in the FDIL scenario. A simple way to address this problem is to train different models for each incremental task. However, this requires extensive storage and time-consuming re-training. Additionally, all these models need to be saved for future use which is laborious because we need to manually select the appropriate mode for each inference data. Instead of training a separate model for each different task, a better solution is to ideally have only a single model that can work in all incremental tasks. However, naively training a model for all tasks can lead to terrible catastrophic forgetting when the unique knowledge in the new task overlaps the ones in the previous tasks. 

Unlike in the FCIL scenario, where incremental tasks involve data from different categories, In the FDIL scenario, the incremental tasks involve data from different domains, i.e., handwritten digit data and cartoon digit data. Compared to the data from different categories in FCIL. i.e., cat data and dog data, there is more shared feature information between the different incremental tasks, which can facilitate the overall FDIL task\cite{liu2019compact}.
The key to efficiently learning new tasks while memorizing previous tasks in the FDIL scenario is to identify the correlation between incremental tasks and then share the useful common knowledge in different tasks. Like \cite{yoon2021federated} in FCIL, we consider each client maintains their personalized models while the number of personalized models is under a given threshold due to the limited capacity of each client. Given the correlation between tasks, each client can maintain several personalized models and share the same model between similar tasks. In this way, the client can choose a model with tasks most similar to the new task, enabling the acquisition of new task knowledge while retaining existing task knowledge. This dual benefit permits the new task to also capitalize on previous knowledge, ultimately enhancing performance. Nonetheless, considering the tasks come incrementally, identifying the inter-task correlations and subsequently training the selected model is challenging. 

To tackle these limitations, we in this paper propose a novel \textbf{P}ersonalized \textbf{Fed}erated \textbf{D}omain-\textbf{I}ncremental \textbf{L}earning (pFedDIL) approach which adaptively matches and migrates knowledge among different incremental tasks. 
More specifically, we propose that each client employs a small auxiliary classifier for each personalized model which is trained to distinguish itself from other tasks. Using the auxiliary classifiers of previous tasks, each client can adaptively identify the correlation of the new task to all previous tasks. Based on these correlations, each client can match the new task with a suitable model, i.e., a new initialized model or a previous model but with similar knowledge. To maximize the training efficiency of the matched model over the new task, we further allow each client to migrate knowledge from other not matched but similar personalized models in a weighted manner, where the weights are the correlations obtained previously. 
Furthermore, to reduce the size of auxiliary classifiers while enhancing their performance, we propose sharing partial parameters between the target personalized model and the auxiliary classifier.
Finally, considering the presence of multiple personalized local models within each client that could lead to additional inference costs, we apply a weighted ensemble distillation over these models where the weights are also obtained from the output of the auxiliary classifiers.

Through extensive experiments on several datasets and different settings, we show that the proposed pFedDIL significantly improves the model accuracy as compared to state-of-the-art approaches. The major contributions of this paper are summarized as follows:
\begin{itemize}
    \item We are the first to formulate the problem of federated domain-incremental learning. To address this issue, we propose a novel method named pFedDIL which enables each client to match and migrate knowledge from previous tasks to alleviate catastrophic forgetting.

    \item To adaptively employ incremental task learning strategy and migrate knowledge, we define the knowledge matching intensity by the auxiliary classifier to discern the correlation between the new task and previous tasks.
    
    \item To facilitate the adaptive selection of inference model among a set of personalized local models, we coordinate all personalized local models to predict weighted by normalized output by auxiliary classifiers. Besides, we further propose sharing partial parameters between the auxiliary classifier and the target classification model to condense model parameters.

    \item We carry out extensive experiments on various datasets and different settings. Experimental results illustrate that our proposed model outperforms the state-of-the-art methods by up to 14.35\% in terms of average accuracy on all tasks.  
\end{itemize}

\section{Related Work}
\textbf{Federated Learning.}
Federated learning (FL) is a distributed learning approach that trains a global model by merging multiple locally trained models on private datasets\cite{li2020review}. 
A significant challenge in FL is data heterogeneity, where data distribution among clients is non-identically and independently distributed (Non-IID), negatively impacting model performance\cite{Jeong2018CommunicationEfficientOM,wang2023fedcda, Shen2020FederatedML}. 
FedAvg is a notable FL method that refines the global model by incorporating parameters trained on local private data\cite{mcmahan2017communication}.
\cite{li2020federated} propose a proximal term to align closely with the global model. 
The federated distillation method\cite{wang2023dafkd} suggests distilling multiple local models into a global one by only aggregating soft output predictions. 
However, existing FL approaches typically assume that the data in each client is static, which cannot account for incremental data with domain shift. 

\noindent\textbf{Incremental Learning.}
Incremental learning (IL) is a learning paradigm that empowers a model to learn sequentially from a series of tasks while preserving the knowledge acquired from previous tasks\cite{wang2024comprehensive,zhao2023does}. IL can be segmented into three primary variants: task-incremental learning\cite{Maltoni2018ContinuousLI}, class-incremental learning\cite{rebuffi2017icarl} and domain-incremental learning\cite{mirza2022efficient}. Existing strategies can further be grouped into three categories: replay-based methods\cite{rolnick2019experience}, regularization-based methods\cite{Jung2020ContinualLW} and parameter isolation methods\cite{chen2020mitigating}.
Most IL methods are designed for task-IL or class-IL and are typically server-based. This work focuses on the federated domain-incremental learning scenario and proposes the pFedDIL technique.

\begin{figure*}[t]
\centering
  \includegraphics[width=0.75\linewidth]{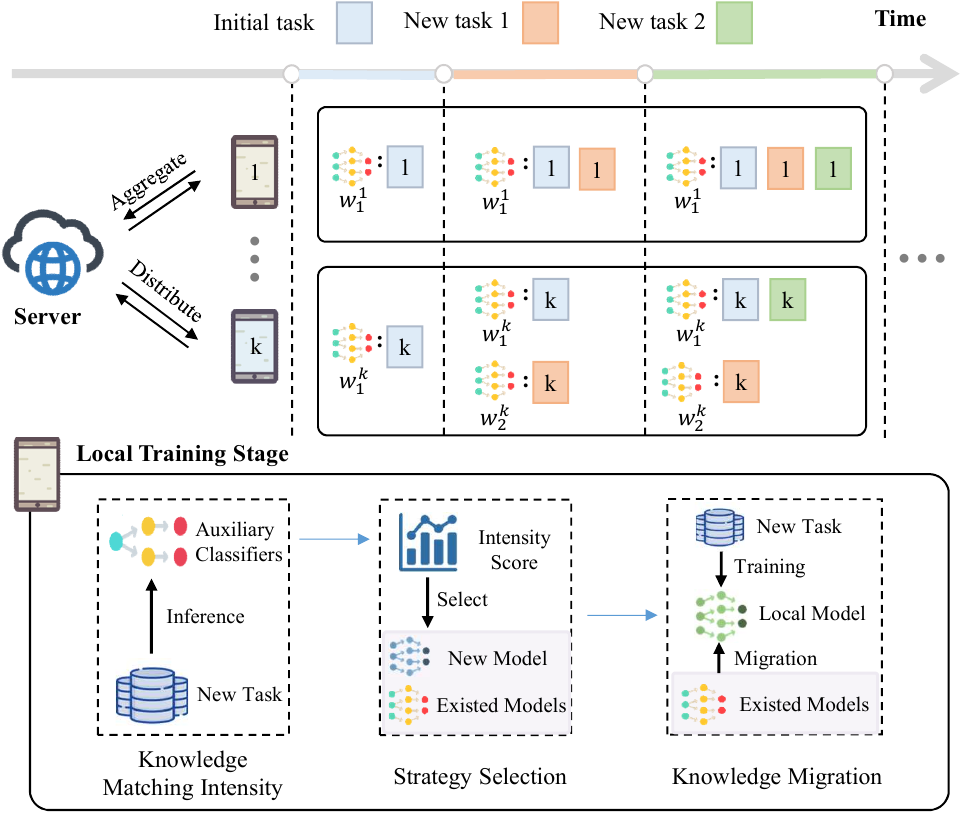}
  \caption{Illustration of the pFedDIL framework. In the training stage, each client randomly receives one domain dataset without replacement to form a task, where different colors represent different domains. Clients can adaptively employ an appropriate incremental task learning strategy and migrate the knowledge from previous tasks with the knowledge matching intensity.}
  \label{framework}
\end{figure*}

\noindent\textbf{Federated Incremental Learning.}
Federated Incremental Learning (FIL), an approach that targets the dynamic adaptation of streaming tasks within each client's environment, emphasizes the seamless integration of new data into the global model while preserving the acquired knowledge from previous data streams. Despite its crucial role, FIL has garnered attention relatively recently, with \cite{yoon2021federated} pioneering the field by focusing on Task-Incremental Learning (Task-IL) in FL. In parallel, studies such as \cite{yu2024personalized} takes the multi-granularity expression of knowledge into account, promoting the spatial-temporal integration of knowledge in FCIL. Conversely, \cite{Dong_2023_ICCV,10323204} simplify the problem's complexity by presupposing clients possess sufficient storage capacity to retain historical examples and engage in data sharing, a scenario that deviates from the conventional federated learning paradigm. Beyond image classification, investigations like \cite{Jiang_2021,10204005} extend the horizons of FIL into diverse domains. Our research focuses on the domain shift between different tasks and the transfer of the common knowledge.

\section{Methodology}
We first formulate our problem definition and then present our pFedDIL approach. Figure \ref{framework} gives an overview of our method.

\subsection{Problem Definition}
In incremental learning, each model learns from a sequence of streaming tasks \{$\mathcal{T}_{(1)},\mathcal{T}_{(2)}\\,\ldots\mathcal{T}_{(n)}$\} where $\mathcal{T}_{(t)}$ denotes the dataset of the $t$-th task; $\mathcal{T}_{(t)} = \{x_i^t,y_i^t\}_{i=1}^{N^t}$, which has $N^t$ pairs of sample data $x_i^t \in X^t$ and corresponding label $y_i^t \in Y^t$. $X^t$ and $Y^t$ represent the domain space and label space for the $t$-th task. In this paper, we focus on the variant of domain incremental learning, where all tasks share the same numbers of classes i.e, $\mathcal{Y}^{1} = \mathcal{Y}^{t}, \forall t \in [1,n]$. As the sequence of tasks arrives, the client needs to learn the new task while their domain and data distribution changes, i.e, $\mathcal{X}^{1} \neq \mathcal{X}^{t}, \forall t \in [1,n]$.

We consider the federated domain-incremental learning scenario. We aim to train a global model for $K$ total clients in the federated domain-incremental learning setting. We consider the local client $k$ can only access the local private streaming tasks \{$\mathcal{T}^k_{(1)},\mathcal{T}^k_{(2)},\dots,\mathcal{T}^k_{(n)}$\}. Now our goal is to train a set of personalized local models $\widetilde{w}^k = [w^k_1,w^k_2,\ldots,w^k_d]$ for each client $k$ over all existing $t$ tasks $\mathcal{T} = \sum_{n=1}^t\sum_{k=1}^K \mathcal{T}^k_{(n)}$. Here each personalized local model corresponds to at least one task, and possibly multiple tasks (i.e. client $k$ may train the task $\mathcal{T}^k_{(1)}$ and $\mathcal{T}^k_{(2)}$ with the model $w^k_1$). Then, we can formulate the goal as :
\begin{align}\label{parameterAvg}
\mathop{\arg\min}_{\widetilde{w}} \mathcal{L}(\widetilde{w}) = \sum_{n=1}^t\sum_{k=1}^K \frac{1}{|\mathcal{T}|} \mathcal{L}_{(n)}^k(\widetilde{w}). 
\end{align}
where $\mathcal{L}_{(n)}^k(\widetilde{w}) = \mathbb{E}_{(x,y)\sim \mathcal{T}^k_{(n)}}[l(\widetilde{w};(x,y))]$ is the empirical loss of client $k$ on $n$-th task with the personalized local model.

\subsection{The Proposed pFedDIL Approach}
The key idea of the pFedDIL is to identify the correlation between the new and previous tasks such that clients can adaptively migrate knowledge from previous tasks to improve performance. 
Specifically, in each communication round, the client in pFedDIL trains the local model with its private sequence of tasks and the server aggregates local models from all participating clients. Then, when it comes to new tasks, each client first employs the existing personalized models' auxiliary classifiers to calculate the correlations denoted as knowledge matching intensity and applies the appropriate incremental task learning strategy: training the new task with a previous personalized local model which has the most similar knowledge or training with a new initial model. Simultaneously, each client can migrate knowledge from other personalized local models with the knowledge matching intensity to to assist in training. During the inference stage, the final inference result is aggregated by all personalized local models weighted by the corresponding auxiliary classifiers. Moreover, we propose sharing partial parameters between the auxiliary classifier and the target personalized model to condense model parameters. The workflow of the pFedDIL is shown in Algorithm \ref{pFedDIL}. 

\noindent\textbf{Knowledge Matching Intensity.}\ To assess the similarity between different incremental tasks, where a higher similarity indicates a greater amount of common knowledge, we propose the knowledge matching intensity. When it comes to the new incremental task: $(t+1)$-th task, each client first needs to compute the knowledge matching intensity between the $(t+1)$-th task and previous $t$ tasks. To achieve the adaptive knowledge matching intensity estimation for each client, we define the auxiliary classifier, a binary classification deep neural network $\theta$ for the corresponding local task $T$, which is trained by the discrimination of the sample belonging to the task $T$. As a new task emerges, the auxiliary classifier can output the similarity between different tasks based on the extracted feature representation. Supposed that the client $k$ has converged on $t$ tasks within $d$ local existing task models when it comes to the incremental $(t+1)$-th task, the client needs to calculate the knowledge matching intensity $\widetilde{\rho}_k = [\rho^k_1,\rho^k_2,\ldots,\rho^k_d]$ with each auxiliary classifier $\widetilde{\theta}^k = [\theta^k_1,\theta^k_2,\ldots,\theta^k_d]$:
\begin{align}\label{intensity}
    \widetilde{\rho}^k = \frac{1}{N^{t+1}}\sum_{i=1}^{N^{t+1}}{f(x^{t+1}_i;\widetilde{\theta}^k)}.
\end{align}
where the $f(\cdot)$ denotes the output of the auxiliary classifier, $\widetilde{\theta}^k$ denotes a set of personalized local auxiliary classifiers in client $k$. Each auxiliary classifier is trained by minimizing the total empirical loss over the corresponding task's local dataset $\mathcal{T}$, which can be formulated as:
\begin{align}
    \mathcal{L}^k(\theta) = \frac{1}{\mathcal{T}} \sum_{i=1}^{\mathcal{T}} \mathcal{L}_{CE}(\theta; x_i, y_i).
    \label{ACLoss}
\end{align}
where $\mathcal{L}^k(\theta)$ is the local loss in the client $k$ and $\mathcal{L}_{CE}$ is the cross-entropy loss function that measures the difference between the prediction and the ground truth labels.

\begin{algorithm}[t]
   \caption{pFedDIL}
   \label{pFedDIL}
   \SetKwInOut{Input}{Input}\SetKwInOut{Output}{Output}
    \Input{$T$: the communication round; $K$: client number; \\ $C$: the
            fraction of active client in each round; \\ $\{\mathcal{T}_{(t)}\}_{t=1}^n$: the distributed dataset with $n$ tasks; \\ $w$: the parameter of the target classification model; \\ $\theta$: the parameter of the auxiliary classifier.}
            
    Initialize the parameter $w$ and $\theta$; 
    
    \For{$t=1$ {\bfseries to} $T$}{
        $S_t \leftarrow $ \text{(random set of}\ $[C \cdot K]$ clients);
        \For{each selected client $k \in S_t$ {\rm\bfseries in parallel}}{
        receives $w_t$ from the server; \\
        calculate knowledge matching intensity $\widetilde{\rho}^k$ with (\ref{intensity});\\
        select learning strategy $w^k_{t+1}$ with (\ref{ILselect});\\
         set local models $w_{t+1}^k$ and $\theta^k_{t+1}$; \\
        \For{$e=1$ {\rm\bfseries to} $E$}{
            update $\theta^k_{t+1}$ with (\ref{ACLoss}); \\
            update $w_{t+1}^k$ through adaptive knowledge migration with (\ref{localloss});
        }
        pushes $w_{t+1}^k$ to the server. \\
        }
        $w_t \leftarrow \text{ServerAggregation}(\{w^k\}_{k \in S_t})$
    }
\end{algorithm}

\noindent\textbf{Incremental Task Learning Strategy.}\ After acquiring the knowledge matching intensity $\widetilde{\rho}_k$ with equation (\ref{intensity}), the client can select the appropriate incremental task learning strategy. An intuition is that the client will prefer to train with previous personalized local models when the knowledge matching intensity exceeds the $\lambda$, which means the high-level similarity between the new task and the previous tasks so that the model can converge on the new task with slight adjustments. Besides, the client will sacrifice some storage by training the new task with the new initial model parameters to ensure accuracy for both the new task and previous tasks. Here we define the incremental task learning strategy as follows:
\begin{equation}\label{ILselect}
w^k_{t+1} = OPT
\begin{cases}
\hat{w}& \text{ $ if \ max(\widetilde{\rho}_k) < \lambda $ } \\
\widetilde{w}^k[m]& \text{ $ else \ max(\widetilde{\rho}_k) = \rho_k^m \geq \lambda$ }.
\end{cases}
\end{equation}
where $\lambda$ is a hyper-parameter, $\widetilde{w}^k = [w^k_1,w^k_2,\ldots,w^k_d]$, $\hat{w}$ denotes new initial model parameters, and $OPT$ means the alternative of existed personalized local models and a new initial model. 

\noindent\textbf{Knowledge Migration.}\ Then, during the training of $(t+1)$-th task, each client respectively migrates the knowledge from other existing personalized local models to assist the training with the calculated knowledge matching intensity $\widetilde{\rho}_k$, minimize the following objective:
\begin{alignat}{2}\label{localloss}
    &\min_{w^k_{t+1}} \mathcal{L}^k_{Local}(w^k_{t+1}) &&= \mathcal{L}^k_{(t+1)}(w^k_{t+1}) + \mathcal{L}^k_{KM}(w^k_{t+1}),  \nonumber \\
    &\text{where} \quad \mathcal{L}^k_{KM}(w^k_{t+1}) &&= \widetilde{\rho}_k \cdot ||w^k_{t+1}-\widetilde{w}^k||^2, \nonumber \\
    &&&= \sum_{i=1}^d\rho^i\cdot||w^k_{t+1}-w^k_i||^2.
\end{alignat}
where $\mathcal{L}^k_{t+1}(w^k_{t+1})$ is defined in (\ref{parameterAvg}) and $\mathcal{L}^k_{KM}(w^k_{t+1})$ refers to the Knowledge Migration (KM) loss.

\noindent\textbf{Inference Stage.} During the inference stage, taking into account that each client may have multiple personalized models, how to automatically select the appropriate model for inference is an urgent problem to solve. We have observed that ensemble learning, which enhances model performance by integrating knowledge from multiple smaller models, has achieved widespread success\cite{dong2020survey}. Utilizing the calculated knowledge matching intensity as the weight for the personalized model inference results, we achieve automated ensemble inference for the inference data. For every inference sample $x \in \hat{\mathcal{T}}$, the client $k$ computes the weights of each personalized local models $\widetilde{w}^k$ with the auxiliary $\widetilde{\theta}^k$ as $f(\widetilde{\theta}^k;x)$ and then normalizes it into the probability:
\begin{align}\label{eq:Importance}
        \hat{\alpha^k_i} = \frac{f(\theta^k_i;x)}{\sum_{i=1}^{d}f(\theta^k_i;x)}.
\end{align}
which guarantees that $\sum_{i=1}^{d}\hat{\alpha}^k_i=1$. Finally, the client inputs the inference sample $x$ into each personalized local models $\widetilde{w}^k$ to obtain the soft predictions $s(\widetilde{w}^k; x)$ and final weighed prediction:
\begin{align}\label{prediction}
       Inference(\widetilde{w}^k;x) = \sum_{i=1}^{d} \hat{\alpha^k_i} \cdot s(w^k_i; x)).
\end{align}

\noindent\textbf{Partial Parameters Sharing.} Motivated by the idea of multi-task learning where sharing the encoder between different tasks can promote each other\cite{standley2020tasks}, we propose sharing partial parameters between the auxiliary classifier $\theta^k$ and the target classification model $w^k$ to solve this problem. The intuition behind this is that both the auxiliary classifier and the classification model have to distinguish the sample from the extracted features. Besides, another benefit of sharing layers is that the model parameters can be reduced.

To this end, we propose sharing the front model layers between the two models which are used to extract features, as illustrated in Figure~\ref{framework}. Specifically, by denoting the $c$-layer shared extractor by $\tilde{w}^{k}=[\tilde{w}^{k,1},\ldots,\tilde{w}^{k,c}]$ where $\tilde{w}^{k,i}$ is the $i$-th layer of the shared extractor, we can re-write the $n_d$-layer auxiliary classifier by $\theta^k=[\tilde{w}^{k}, \theta^{k,c+1},\ldots,\theta^{k,n_d}]$, and the $n$-layer target classification model $w^k$ by $w^k=[\tilde{w}^{k},w^{k,c+1},\ldots,w^{k,n}]$ with $w^{k,i}$ as the $i$-th layer of the classification model $w^{k,i}$. With the shared layers, the discriminator and the classification model are trained jointly:
\begin{align}\label{eq:SharedLayers}
        \min_{\theta^{k}, w^{k}} \mathcal{L}_J^k(\theta^k, w^{k})= \mathcal{L}^k(\theta^{k})+\mathcal{L}^{k}(w^{k}).
\end{align}
where $\mathcal{L}^k(\theta^{k})$ refers to (\ref{ACLoss}) and $\mathcal{L}^{k}(w^{k})$ is defined in (\ref{parameterAvg}).

\begin{table}[t]
    \centering
        \caption{Performance comparison by Top-1 test accuracy for our model and baseline methods on Digit-10 (data heterogeneity: $\alpha$ = 0.1).}
    \renewcommand\arraystretch{1.2}
      \resizebox{0.85\linewidth}{!}{
    \begin{tabular}{c |c c c c c c c| c |c}
    \toprule[1pt]
    & \multicolumn{9}{c}{\textbf{Digit-10} ($\lambda$ = 0.5)}\\
    \hline
         Method &MNIST &$\rightarrow$ & EMNIST &$\rightarrow$ & USPS&$\rightarrow$  & SVHN & Avg &$\Delta(\uparrow)$\\
       \hline
       FedAvg\cite{mcmahan2017communication}&92.82&$\rightarrow$&84.02&$\rightarrow$&81.02&$\rightarrow$&71.87&82.43&6.85$ \uparrow$\\
       FedProx\cite{li2020federated}&93.01&$\rightarrow$&84.68&$\rightarrow$&78.30&$\rightarrow$&76.42 &83.10&  6.18$\uparrow$\\
      FedCIL\cite{qi2023better}&94.64&$\rightarrow$
       &87.52&$\rightarrow$&82.16&$\rightarrow$&80.92 &86.31&2.97$\uparrow$\\     
       DANN\cite{ganin2016domain} + FL&\textbf{96.07}&$\rightarrow$ &86.71&$\rightarrow$&79.11&$\rightarrow$&72.14 &83.51& 5.77$\uparrow$\\
       Source-Only&92.82&$\rightarrow$&82.15&$\rightarrow$&75.53&$\rightarrow$&69.06& 79.89&9.39$\uparrow$\\
       Sharing&92.67&$\rightarrow$&82.91&$\rightarrow$&76.17&$\rightarrow$ &71.96&80.93&8.35$\uparrow$\\
        \hline

       pFedDIL -w/o Migration &92.82&$\rightarrow$ &87.09&$\rightarrow$&84.00&$\rightarrow$&81.89&86.45 &2.83$\uparrow$\\
       pFedDIL -w/o Sharing &92.79&$\rightarrow$ &90.13&$\rightarrow$&\textbf{87.59}&$\rightarrow$&86.19&89.18&0.11$\uparrow$\\
       pFedDIL -w/o Correlation &92.82&$\rightarrow$ &89.06&$\rightarrow$&84.72&$\rightarrow$&84.30&87.73 &1.55$\uparrow$\\

       \hline
       pFedDIL &92.82&$\rightarrow$ &\textbf{90.39}&$\rightarrow$&87.14&$\rightarrow$&\textbf{86.75}& \textbf{89.28}&/\\
       \hline
       Upper Bound -Disjoint&92.82&$\rightarrow$&90.33&$\rightarrow$&93.41&$\rightarrow$&87.97&91.13&1.85$\downarrow$\\
    \bottomrule[1pt]
    \end{tabular}}
    \label{digit}
\end{table}
\begin{table}[t]
    \centering
        \caption{Performance comparison by Top-1 test accuracy for our model and baseline methods on Office-31 (data heterogeneity: $\alpha$ = 1).}
    \renewcommand\arraystretch{1.2}
      \resizebox{0.75\linewidth}{!}{
    \begin{tabular}{c |c c c c c| c |c}
    \toprule[1pt]
    &\multicolumn{7}{c}{\textbf{Office-31} ($\lambda$ = 0.8)}\\
    \hline
         Method & Amazon&$\rightarrow$ &Dlsr&$\rightarrow$&Webcam&Avg &$\Delta(\uparrow)$\\
       \hline
       FedAvg\cite{mcmahan2017communication}&46.53&$\rightarrow$&24.17&$\rightarrow$&29.34&33.35&  12.82$\uparrow$\\
       FedProx\cite{li2020federated}&45.19&$\rightarrow$& 25.68&$\rightarrow$&32.23&34.37&11.80$\uparrow$\\
      FedCIL\cite{qi2023better}&49.38&$\rightarrow$&39.65&$\rightarrow$ &44.04&44.36&1.81$\uparrow$\\     
       DANN\cite{ganin2016domain} + FL&\textbf{51.97}&$\rightarrow$&35.96&$\rightarrow$ &43.08&43.67& 2.50$\uparrow$\\
       Source-Only&46.53&$\rightarrow$&20.61&$\rightarrow$&28.32&31.82&14.35$\uparrow$\\
       Sharing&46.11&$\rightarrow$&24.23&$\rightarrow$&34.25&34.86&11.31$\uparrow$\\
        \hline

       pFedDIL -w/o Migration&46.53&$\rightarrow$&40.17&$\rightarrow$ &44.55&43.75&2.42$\uparrow$\\
       pFedDIL -w/o Sharing &46.63&$\rightarrow$&42.90&$\rightarrow$ &47.03&45.52&0.65$\uparrow$\\
       pFedDIL -w/o Correlation &46.53&$\rightarrow$&41.66&$\rightarrow$ &47.19&45.13&1.04$\uparrow$\\

       \hline
       pFedDIL &46.53&$\rightarrow$&\textbf{43.87}&$\rightarrow$&\textbf{48.12}&\textbf{46.17}&/\\
       \hline
       Upper Bound -Disjoint&46.53&$\rightarrow$&50.17&$\rightarrow$&56.21&50.97&4.80$\downarrow$\\
    \bottomrule[1pt]
    \end{tabular}}
    \label{office}
\end{table}

\begin{table}[t]
    \centering
        \caption{Performance comparison by Top-10 test accuracy for our model and baseline methods on DomainNet (data heterogeneity: $\alpha$ = 10)}
    \renewcommand\arraystretch{1.2}
      \resizebox{\linewidth}{!}{
    \begin{tabular}{c |c c c c c c c c c c c| c |c}
    \toprule[1pt]
    &\multicolumn{13}{c}{\textbf{DomainNet} ($\lambda$ = 0.8)}\\
    \hline
         Method &Clipart&$\rightarrow$  & Infograph&$\rightarrow$  & Painting&$\rightarrow$&Quickdraw&$\rightarrow$&Real&$\rightarrow$&Sketch&Avg &$\Delta(\uparrow)$\\
       \hline
       FedAvg\cite{mcmahan2017communication}&48.43&$\rightarrow$&37.18&$\rightarrow$&45.80&$\rightarrow$&45.32&$\rightarrow$&44.91&$\rightarrow$&50.25&45.32&6.38$\uparrow$\\
        FedProx\cite{li2020federated}&47.39&$\rightarrow$&38.43&$\rightarrow$&44.31&$\rightarrow$&47.96&$\rightarrow$&42.38&$\rightarrow$&51.77&45.37&6.32$\uparrow$\\
       FedCIL\cite{qi2023better}&\textbf{52.14}&$\rightarrow$
       &43.68&$\rightarrow$&47.10&$\rightarrow$&48.75&$\rightarrow$ &42.89&$\rightarrow$&51.26&47.64&4.05$\uparrow$\\
       DANN\cite{ganin2016domain} + FL&50.07&$\rightarrow$&39.74&$\rightarrow$  &43.73&$\rightarrow$&45.08&$\rightarrow$& 43.28&$\rightarrow$&52.96&45.81&5.88$\uparrow$\\
       Source-Only&48.43&$\rightarrow$&31.01&$\rightarrow$&33.12&$\rightarrow$ &38.15&$\rightarrow$&37.62&$\rightarrow$&42.85&38.53&13.16$\uparrow$\\
       Sharing& 48.43&$\rightarrow$&36.03&$\rightarrow$&39.58&$\rightarrow$& 41.80&$\rightarrow$&40.41&$\rightarrow$&44.76&41.84&9.86 $\uparrow$\\
  
        \hline
       pFedDIL -w/o Migration &48.43&$\rightarrow$ & 43.83&$\rightarrow$   &50.05&$\rightarrow$& 49.44&$\rightarrow$&46.10&$\rightarrow$ &54.90& 48.79&2.90$\uparrow$\\
       pFedDIL -w/o Sharing &48.52&$\rightarrow$& \textbf{47.37}&$\rightarrow$& 51.16&$\rightarrow$&53.98&$\rightarrow$&50.11&$\rightarrow$&55.24&51.06&0.63$\uparrow$\\
       pFedDIL -w/o Correlation&48.43&$\rightarrow$&46.71&$\rightarrow$    &51.09&$\rightarrow$&52.34&$\rightarrow$&48.07&$\rightarrow$ &56.95&50.60&1.09$\uparrow$\\

       \hline
       pFedDIL&48.43&$\rightarrow$ &47.19&$\rightarrow$&\textbf{52.48}&$\rightarrow$  &\textbf{54.07}&$\rightarrow$&\textbf{50.36}&$\rightarrow$&\textbf{57.61}&\textbf{51.69}&/ \\
       \hline
       Upper Bound -Disjoint&48.43&$\rightarrow$ &52.33&$\rightarrow$& 54.21&$\rightarrow$&52.89&$\rightarrow$&53.62&$\rightarrow$&58.06&53.26&1.57$\downarrow$\\
    \bottomrule[1pt]
    \end{tabular}}
    \label{domainnet}
\end{table}

\section{Experiments}
\subsection{Setup}
\noindent\textbf{Dataset:} We conduct all our experiments with heterogeneous dataset partition on three datasets. In FDIL scenarios, new domains are introduced gradually. The dataset initially contains samples from a specific domain, and new domains are introduced at later stages, enabling models to adapt and generalize to new unseen domains.
\begin{itemize}
    \item \textbf{Digit10:} Digit-10 dataset contains 10 digit categories in four domains: \textbf{MNIST}\cite{lecun2010mnist}, \textbf{EMNIST}\cite{cohen2017emnist}, \textbf{USPS}\cite{hull1994database}, \textbf{SVHN}\cite{netzer2011reading}.Each dataset is a digit image classification dataset of 10 classes in a specific domain, such as handwriting style.
    
    \item \textbf{Office31:\cite{saenko2010adapting}} A dataset with images from three different domains: Amazon, Webcam, and DSLR. It consists of 31 object categories, with each domain having around 4,100 images.

    \item \textbf{DomainNet:\cite{peng2019moment}} A large-scale dataset with images from six different domains: Clipart, Painting, Real, Sketch, Quickdraw, and Infograph. It contains over 0.6 million images across 345 categories.
\end{itemize}


\noindent\textbf{Baseline:} For a fair comparison with other key related works, we follow the same protocols proposed by \cite{mcmahan2017communication} to set federated learning(FL) task and utilize the domain-incremental learning (DIL) task by \cite{mirza2022efficient}. We evaluate all baselines under the above settings, including two representative FL models \textbf{FedAvg\cite{mcmahan2017communication}} and \textbf{FedProx\cite{li2020federated}}, which is better at tackling heterogeneity in federated networks than FedAvg, and the following several methods in FDIL scenario:

\noindent\textbf{FedCIL\cite{qi2023better}:} This approach employs the ACGAN backbone to generate synthetic samples to consolidate the global model and align sample features in the output layer. Authors conduct experiments in FCIL scenarios, and here we adopt it to the FDIL setting.

\noindent\textbf{Domain Adaptation + FL:} Here we adopt the robust adversarial-based method DANN\cite{ganin2016domain}. This baseline mainly follows the domain adaptation paradigm which is different from the incremental learning and is often prone to issues like catastrophic forgetting.

\noindent\textbf{Source-Only:} Here we maintain the initial model trained on the initial domain data completely fixed throughout all tasks. This baseline does not perform any adaptation, modification or replacement to its parameters and serves as a lower bound.

\noindent\textbf{Sharing:} Inspired by the multi-task learning scenario\cite{ruder2017overview}, we adopt all front layers before the last fully connected layer as shared layers, and we use relevant different fully-connected layers to get outputs of different domains. During inference, the fully-connected layer that matches the corresponding domain ID will be applied.

\noindent\textbf{Disjoint:} Here we train a separate model for each domain. During inference, the model which matches the corresponding domain ID is used. This baseline requires additional space to store parameters and additional cost to obtain inference-task ID. It is a degradation of domain-incremental learning and provides an upper bound here.

\noindent\textbf{Configuration:} Unless otherwise mentioned, we set the number of local training epoch \emph{E} = 20, communication round \emph{T} = 180 for each domain. Here we train each task until convergence before the arrival of a new task with communication rounds $T$. For local training, the batch size is 32, the learning rate for our models is 0.001 and the weight decay is $1e-3$. Like most FL methods on Non-IID data distribution\cite{mcmahan2017communication}, we use the Dirichlet distribution ($\alpha$) to simulate the data heterogeneity from all domains and we apply the same data heterogeneity for all domains in the same dataset. For the classifier in all methods, we employ ResNet18\cite{he2016deep} as the basic backbone. For the multi-task learning structure in our approach, we treat all previous layers before the last fully connected layer as share layers, and we use two different fully connected layers to get outputs as the auxiliary classifier result and target classification result. The total clients' number is 20 for Digit-10 and DomainNet with an active ratio \emph{k} = 0.4, 10 for Office-31 with an active ratio \emph{k} = 0.4. 

\subsection{Performance Overview}
\noindent\textbf{Test Accuracy.} Table \ref{digit},\ref{office},\ref{domainnet} shows the performance of our pFedDIL and other baseline methods on three datasets, where "$\Delta$" denotes the improvement of our method with other baselines. We carry out experiments against data heterogeneity on each dataset, and pFedDIL achieves the best performance in most cases by a margin of 1.81\%$\sim$14.35\% in terms of average accuracy. Due to our challenging FDIL settings, there is a notable gap between all baselines and the upper performance bound. However, pFedDIL almost achieves comparable results with less storage and adaptive task-id inference during the inference stage. These results demonstrate that the appropriate incremental task learning strategy and adaptive knowledge migration technique in pFedDIL can alleviate catastrophic forgetting and thus improve performance.

\begin{figure*}[h]
  \centering
    \includegraphics[width=\textwidth]{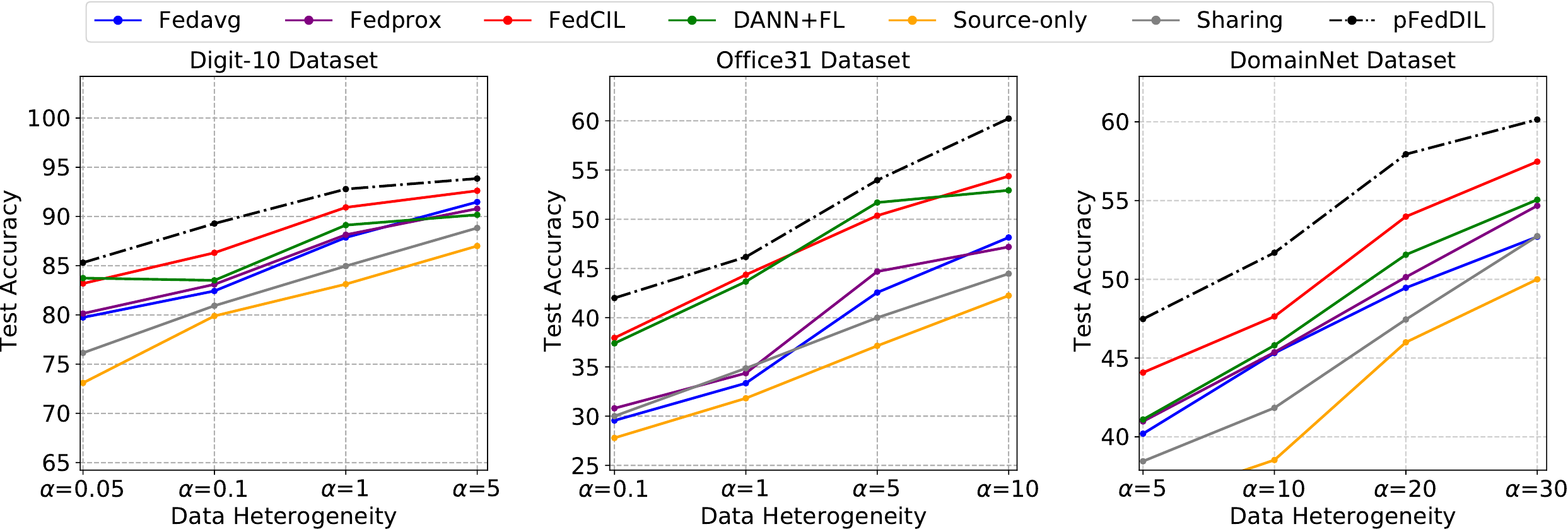}
  \caption{Visualized performance w.r.t data heterogeneity for three datasets. }
  \label{data_heterogeneity}
\end{figure*}
\noindent\textbf{Ablation Study.} 
As shown in Tables \ref{digit},\ref{office},\ref{domainnet}, we evaluate the effects of each module in our model via ablation studies. pFedDIL -w/o Migration, pFedDIL -w/o Sharing and pFedDIL- w/o Correlation denote the performance of our model without using knowledge migration, parameters sharing and correlation factor in the inference stage. Compared with pFedDIL, the performance of pFedDIL -w/o Migration, pFedDIL -w/o Sharing and pFedDIL- w/o Correlation degrades evidently with a range of 0.11\%$\sim$2.90\%. Specifically, the adaptive knowledge migration technique in pFedDIL plays a much more important role with the obvious improvement in test accuracy, and the partial parameters sharing trick condenses model parameters by sharing the feature extracted layer while not causing a remarkable impact on test accuracy. Experiment results verify the effectiveness of all modules, confirming all modules are essential to train a robust federated domain-incremental model.

\noindent\textbf{Data heterogeneity.}
Increasing data heterogeneity levels in experiments is significant because it allows for a more realistic evaluation of the system's performance in diverse real-world scenarios and assesses its robustness and generalization capabilities. In particular, in the FDIL scenario, incremental data is less likely to appear in the form of IID data. Therefore, it is essential to simulate different levels of Non-IID data to validate the effectiveness of our method in the FDIL scenario. Figure \ref{data_heterogeneity} displays the accuracy with different levels of data heterogeneity on three datasets. As shown in this figure, all methods achieve an improvement in test accuracy with the decline of data heterogeneity. Most notably, pFedDIL always maintains a leading advantage with different levels of data heterogeneity.

\begin{table}[]
\renewcommand\arraystretch{1.2}
    \caption{Evaluation of various methods in terms of the communication rounds. We report the sum of communication rounds required to achieve the best performance on each task and the running time of a training communication round.}
    \centering
    \resizebox{0.95\linewidth}{!}{
    \begin{tabular}{ c  c c c c c c c c c}
    \toprule[1pt]
     \multirow{2}{*}{Dataset} & \multicolumn{2}{c}{FedAvg} & \multicolumn{2}{c}{FedCIL} & \multicolumn{2}{c}{DANN+FL} & \multicolumn{2}{c}{pFedDIL} \\ \cline{2-9}
     & Rounds & Time & Rounds & Time & Rounds & Time & Rounds & Time \\
     \hline
      Digit10 & 480(r) & \textbf{32.14}(s) & 492(r) & 50.06(s) & 553(r) & 46.77(s) & \textbf{379}(r) & \underline{34.91}(s)\\
      Office31 & 463(r) & \textbf{44.96}(s) & 482(r) & 78.32(s) & 488(r) & 62.89(s) & \textbf{411}(r) & \underline{47.58}(s)\\
    DomainNet & 835(r) & \textbf{92.22}(s) & 904(r) & 165.41(s) & 861(r) & 158.01(s) & \textbf{725}(r) & \underline{105.38}(s)\\
    \bottomrule[1pt]
    \end{tabular}}
    \label{round}
\end{table}

\noindent\textbf{Computational complexity and communication cost.}
We treat all previous layers in ResNet18 before the last fully connected layer as shared layers, and we use two different fully connected layers to get outputs as the auxiliary classifier result and target classification result. Compared with baseline methods, we only train one additional fully connected layer as the auxiliary classifier for the binary classification task - the discrimination of the sample belonging to the current task. The training time cost (each communication round) is shown in Table \ref{round}. Therefore, the training of the auxiliary classifier will not increase the overhead or computational complexity significantly and hardly affect the training efficiency.

Table \ref{round} shows the evaluation of various methods in terms of the communication rounds to reach the test accuracy reported in Table \ref{digit}, \ref{office} and \ref{domainnet} (180 communication rounds each task). Here we show the sum of communication rounds required to achieve the performance on each task. pFedDIL {requires the least communication rounds to achieve the reported test accuracy} on all datasets with the knowledge matching. Thus, the local model is easier to converge.

\begin{figure*}[t]
  \centering
  \includegraphics[width=\linewidth]{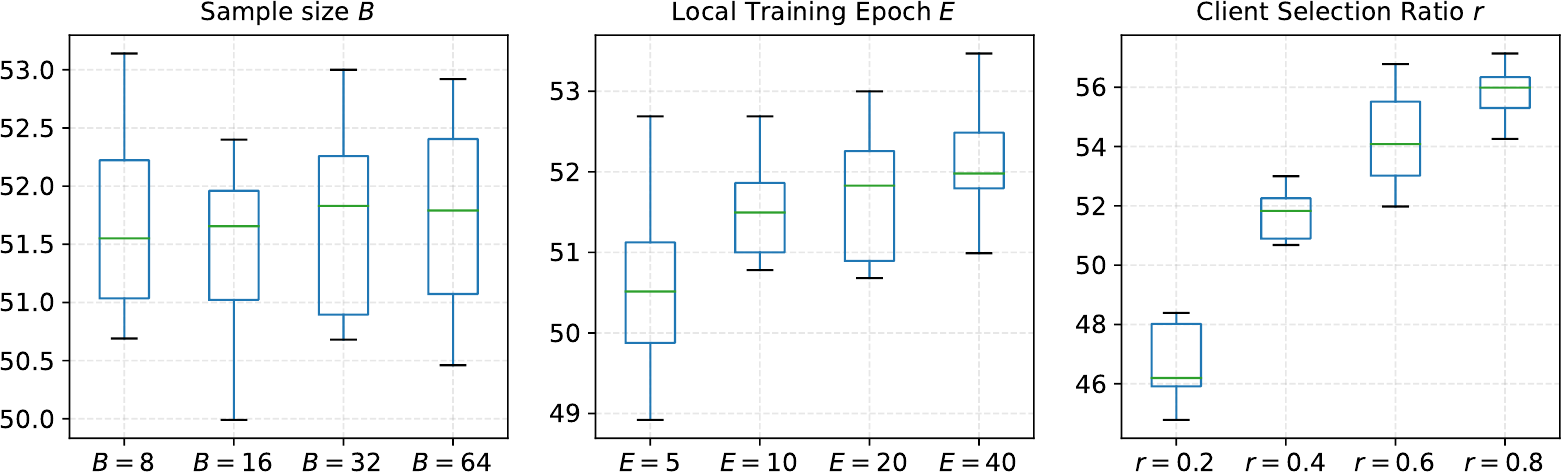}
  \caption{Performance of pFedDIL under different configurations. Here, we select three general parameters in the FL setting: (a) local training epoch $E$, (b) sample size $B$ in the classifier, (c) client selection ratio $r$ of all clients on DomainNet with $\alpha$ = 10.}
  \label{parameter}
\end{figure*}

\noindent\textbf{Parameter sensitivity analysis.}
To figure out whether pFedDIL is sensitive to some specific parameters, we first select some general parameters, which are defined in most FL: local training epoch $E$; sample batch size $B$, and client selection ratio for global aggregation $r$ on DomainNet with $\alpha$ = 10 to carry out experiments. Figure \ref{parameter} shows the performance of pFedDIL under different configurations. pFedDIL achieves better results when we increase the local training epochs at the beginning. However, pFedDIL has a comparable performance when the $E$ is set to 20 and 40. The variance resulting from $E$ and $r$ being less than 0.6 and 13 percentage points respectively is comparable to that of existing works and is not significant. In addition, pFedDIL achieves similar performance with different sample sizes $B$.

\begin{table}[h]
    \caption{Performance of pFedDIL approach on all datasets with different hyper-parameter $\lambda$ values. Here we measure average accuracy and model size over all tasks on each client (each model including the auxiliary classifier costs about 22Mb storage).}
    \centering
    \renewcommand\arraystretch{1.3}
    \resizebox{0.75\textwidth}{!}{
    \begin{tabular}{c c c c c c c}
    \toprule[1pt]
       \multicolumn{2}{c}{Dataset} &  $\lambda$ = 0.0 & $\lambda$ = 0.2 &$\lambda$ = 0.5 &$\lambda$ = 0.8 &$\lambda$ = 1.0 \\
       \hline
       \multirow{2}{*}{Digit-10}& Avg ACC & 82.43&85.02&\textbf{89.28} &90.77 &93.94\\
               & Model Size &22Mb&37Mb&\textbf{46Mb}&71Mb&88Mb\\
        \hline
       \multirow{2}{*}{Office-31}& Avg ACC &33.35 &40.51& 39.98&\textbf{46.17} &50.38\\
               & Model Size &22Mb&42Mb&48Mb&\textbf{53Mb}&66Mb\\
        \hline
       \multirow{2}{*}{DomainNet}&Avg ACC&45.32&48.19&49.65&\textbf{51.69}&53.75\\
               & Model Size &22Mb&73Mb&84Mb&\textbf{91Mb}&132Mb\\
    \bottomrule[1pt]
    \end{tabular}
    }
    \label{lambda}
\end{table}

Then, we conduct more research on the setting of hyper-parameter $\lambda$. When $\lambda$ = 0.0, pFedDIL will degenerate to the normal FedAvg algorithm without any knowledge migration with the single model. When $\lambda$ is set to 1.0, pFedDIL will separately train models for each domain with knowledge migration. As shown in Table \ref{lambda}, we select five different $\lambda$ values to conduct experiments and evaluate the average accuracy and average model size on all personalized local models. In Table \ref{digit},\ref{office}, and \ref{domainnet}, we utilize approximately two auxiliary classifiers per client in Digit-10 and Office-31 and about four in DomainNet. For Table \ref{digit},\ref{office}, we carefully select the optimal value of $\lambda$ for each dataset, striking a balance between accuracy and storage. As shown in Table \ref{lambda}, pFedDIL consistently outperforms other baselines even when $\lambda$ is simply set to 0.5 across all datasets (around two auxiliary classifiers per client in Digit-10 and Office-31 and approximately three in DomainNet). This suggests that a simple parameter setting can outperform existing methods, and with a more refined parameter setting, we can better accommodate practical considerations, achieving an optimal trade-off between accuracy and storage for each client. The \textit{flexibility} of pFedDIL allows users to tune this parameter based on their specific requirements and constraints in real-world scenarios.


\section{Limitations}
Our work is aimed at solving the forgetting problem under the assumption of shared knowledge between tasks in the domain incremental setting. The problem of learning tasks without similar knowledge is rare in real-world scenarios and violates the principle of incremental learning\cite{hadsell2020embracing}. The ultimate goal of incremental learning is to better learn new knowledge with the assistance of previous knowledge while also mastering both previous and new knowledge\cite{wang2024comprehensive}. Furthermore, we investigate the situation where the previous data is strictly unseen to clients, while some research on federated incremental learning assumes that each client can cache some samples for replay\cite{fini2020online}. Both situations exist and there is a must to separately develop appropriate methods. 

\section{Conclusion and Future Work}
In this paper, we seek to tackle the catastrophic forgetting in the FDIL scenario. We propose a personalized federated domain-incremental learning approach based on adaptive knowledge matching, named pFedDIL. We leverage the auxiliary classifier to calculate the knowledge-matching intensity for the incremental task-learning strategy selection and knowledge migration. Furthermore, we propose sharing partial parameters between the target classification model and the auxiliary classifier to condense model parameters. Extensive experiments conducted on various datasets and settings show that our method achieves significant improvement in accuracy.
Although existing works and our method have demonstrated great effectiveness over the FCIL and FDIL scenarios separately, none of them take both scenarios into account. To deploy the FL system in practical settings, it is necessary to simultaneously handle the class incremental tasks and domain incremental tasks. In the future, we seek to move forward in this field.

\section*{Acknowledgements}
This work is supported by National Natural Science Foundation of China under grants 62376103, 62302184, 62206102, Science and Technology Support Program of Hubei Province under grant 2022BAA046.

%
%
\bibliographystyle{splncs04}
\bibliography{main}
\end{document}